\documentclass[runningheads]{llncs}
\usepackage{amssymb}
\usepackage{array}
\usepackage{tabularx}

\usepackage{graphicx}
\usepackage{amsmath}

\DeclareMathOperator*{\argmax}{\arg\!\max}
\DeclareMathOperator*{\argmin}{\arg\!\min}
\newcommand{\citet}{\cite} 
\newcommand{\citep}{\cite}

\usepackage[ruled,vlined]{algorithm2e}
\usepackage{booktabs}
\usepackage[table,xcdraw]{xcolor}
\usepackage{cite}

\begin{document}
%


    \title{Improving Confidence in the Estimation of Values and Norms}
    %
    %
    \author{Luciano Cavalcante Siebert\inst{1} \and
    Rijk Mercuur \inst{2}\and
    Virginia Dignum\inst{2,3} \and \\
    Jeroen van den Hoven \inst{2} \and
    Catholijn Jonker \inst{1,4}
    }
    \authorrunning{L. C. Siebert et al.}
    %
    \institute{Faculty of Electrical Engineering, Mathematics, and Computer Science, \\Delft University of Technology, The Netherlands \\
    \email{l.cavalcantesiebert@tudelft.nl}, ORCID iD: 0000-0002-7531-3154
\\ \and
    Faculty of Technology, Policy and Management, \\Delft University of Technology, The Netherlands \and
    Department of Computing Sciences, Umeå University, Sweden \\ \and
    Leiden Institute of Advance Computer Science, Leiden University, The Netherlands
    }
    \maketitle 
    \begin{abstract}
    Autonomous agents (AA) will increasingly be interacting with us in our daily lives. While we want the benefits attached to AAs, it is essential that their behavior is aligned with our values and norms. Hence, an AA will need to estimate the values and norms of the humans it interacts with, which is not a straightforward task when solely observing an agent's behavior. This paper analyses to what extent an AA is able to estimate the values and norms of a simulated human agent (SHA) based on its actions in the ultimatum game. We present two methods to reduce ambiguity in profiling the SHAs: one based on search space exploration and another based on counterfactual analysis. We found that both methods are able to increase the confidence in estimating human values and norms, but differ in their applicability, the latter being more efficient when the number of interactions with the agent is to be minimized. These insights are useful to improve the alignment of AAs with human values and norms.
    
    \end{abstract}
    
    \section{Introduction}
    As autonomous agents (AAs) become more pervasive in our daily lives, there is a growing need to reduce the risk of undesired impacts on our society \cite{amodei2016concrete,hadfield2017inverse}. Hence, we need design and engineering approaches that consider the implications of ethically relevant decision-making by machines, understanding the AA as part of a socio-technical system \cite[p.48]{dignum2019responsible}. For this, we need to ensure that the decisions and actions made by the AA are aligned with the stakeholders' values (``what one finds important in life'' \cite{van2011ethics}) and norms (what is standard, acceptable or permissible behavior in a group or society \cite{fishbein2011predicting}). 
    
    Values and norms can be seen as criteria for decision-making: values relating to more abstract and context-independent ideals, and norms as more concrete context-dependent rules. Furthermore, since humans use values and norms in their explanations, the use of these concepts allows for a better explanation of the AA \cite{mercuur2019value,malle2006mind,miller2018explanation}. People have a common base system of values and they are relatively stable over the life span of a person \cite{cranefield2017no}, however, values and norms vary significantly for each person and socio-cultural environment, making it difficult or even impossible to write down precise rules describing them. One approach to deal with this variance is to treat aligning the actions of the AA with human values as a learning problem \cite{irving2019ai,soares2017agent}. However, estimating values and norms solely based on observed behavior may lead to ambiguous results. Different relative preferences towards values and norms can bring about the same behavior, i.e. there are many ``reasons why'' that might motivate a given observed behavior (ambiguity problem). In case that an agent's actions are driven by a ``wrong'' set of values and norms, strong ethical consequences may befall. 
    
    This paper studies the conditions under which AAs are able to make confident estimates of one's preference on values and norms from observed behavior. We studied the behavior of simulated agents driven by both values and social norms in the Ultimatum Game (UG), specifically in the proposer role. \cite{nouri2017culture} has used the UG to study the relative importance of values across cultures. We focus on the relative importance of values and norms that guide individual decision making, aiming to provide useful insights to improve the alignment of AAs with human values and norms.
    
    The main contribution of this paper is twofold. First, we propose an agent-based model to play the UG, expanded from  \cite{mercuur2019value}, which uses both values and norms for determining the actions of the agents. Second, we present a method for estimating an agent's relative preferences to a given set of values and, if necessary, to improve the confidence in these estimations. This improvement is realized by interacting with the proposer agent in the UG, either by a free exploration of the search space or by the use of counterfactuals.
    
    The remainder of the paper is structured as follows. The following section presents a brief context about the UG and how we model values and norms in this context. Section 3 presents both the method to estimate the relative preference attributed to an agent's values and norm, and the methods to reduce ambiguity on the estimation. Section 4 presents the main results of this research and section 5 discusses these results. Finally, section 6 presents the conclusions.
    
    \section{Values and norms in the ultimatum game}
    
    \subsection{Scenario: The Ultimatum Game}
    \label{sec:ug}
    We simulate the behavior of the agents in the UG, first introduced by \citep{guth1982experimental}. In the UG, two players negotiate over a fixed amount of money (the `pie'). The proposer demands a portion of the pie, while the remainder is offered to the responder. The responder chooses to accept or reject this proposed split. If the responder chooses to `accept', the proposed split is implemented. If the responder chooses to `reject', both players get no money. The UG thus provides an environment where players have to make decisions about money based on their own judgment.
    
    \vspace{1cm}
    
    The remainder of the paper uses a version of the UG with the following specifics:
    \begin{itemize}
    \item An experiment has 32 players: 16 proposers and 16 responders.
    \item One experiment has 20 rounds. One round in the UG comprises one demand for each proposer and one reply for each matched responder.
    \item The players are paired to a different player each round but do not change roles.
    \item Players are anonymous to each other.
    \item The pie size $P$ is $1000$\footnote{For ease of presentation, we chose to present $P$ with no monetary unit. Empirical work \cite{Oosterbeek2001} shows that the effect of the pie size is relatively small.}.
    \end{itemize}
    \noindent
    These specifics are chosen for pragmatic reasons: the paper uses an empirical dataset based on these specifics \citep{Cooper2011a} and builds upon a model based on these specifics \citep{mercuur2019value}.
    
    This paper models the decision-making of humans in the UG by using values and norms, but there have been many different approaches to it since its first appearance in \cite{guth1982experimental}. One reason for its popularity is that the UG is a classical example of where the canonical game-theoretical agent (i.e., the \textit{homo economics} that only cares about maximizing its direct own welfare) falls short. The \textit{homo economicus} only cares about its direct welfare and thus will accept any positive offer in the UG. Humans, in contrast, reject offers as high as 40\% of the pie \cite{Oosterbeek2001}. Behavioral economics has aimed to explain humans by incorporating learning \cite{Roth1995}, reputation \cite{Fehr2003} or other-regarding preferences. Recently, \cite{mercuur2019value} used values and norms to explain UG behavior. Using a series of agent-based simulation experiments, \cite{mercuur2019value} showed that the model produces aggregate behavior that falls within the 95\% confidence interval wherein human behavior lies more often than the other tested agent models (e.g., a learning \textit{homo economicus} model). Moreover, the model uses concepts that humans use in their explanations. Thus because the model has been shown to reproduce human behavior and uses explainable concepts, the model of \cite{mercuur2019value} provides a starting point to estimate preference towards human values and norms for value alignment.
    
    \subsection{Simulated Human agent (SHA)}
    The simulated human agent (SHA) represents the human in the UG whose values and norms we aim to estimate. The model for the SHA is based on the value-based model and norm-based model presented in \cite{mercuur2019value}. These models focus on the aggregate properties that emerge from pairing these agents in the UG. However, this paper focuses on estimating relative preference towards values and norms of individual agents. Therefore, we are interested in an agent model where one agent uses \emph{both} values and norms. The remainder of this section presents the SHA model in three parts: the value-based agent (from \cite{mercuur2019value}), the norm-based agent (extension of \cite{mercuur2019value}) and an agent that uses both values and norms (new).
    
    \subsubsection{Value-Based Agent} The value-based agent uses the importance an agent attributes to its values to determine what an agent (based on its values) should demand. \cite{mercuur2019value} focuses on two values that are relevant in the UG: wealth and fairness and define $di_a$ as the difference in importance an agent attributes to wealth versus fairness. Based on research by \cite{Schwartz2012} and data on the UG \cite{Cooper2011a}, they then state a number of requirements for a function (see \cite{mercuur2019value} for more details). First, they assume a (perfect) negative correlation between wealth and fairness. Second, valuing wealth is defined as wanting more money. Third, valuing fairness is defined as wanting an equal split of the pie. Fourth, agents should differ in how much money they demand, but not demand below 0.5P.
    The functions that map $di_a$ to the demand $d \in \{0...P\} \subset \mathbb{Z}$ are given by
    \begin{equation}
    \label{func:value}
    valueDemand(di_a) =\argmax_{d\in \{0...P\} \subset \mathbb{Z}} u(d,di_a,P)
    \end{equation}
    and
    \begin{equation}
    \label{func:utility}
    u(d,di_a,P) = -\frac{1.0+0.5di_a}{\frac{d}{P} +0.5} - \frac{1.0-0.5di_a}{\frac{|0.5P-d|}{0.5P}+0.5}
    \end{equation}
    \noindent
    The agent thus calculates the utility for each demand $d$ and returns the demand with the maximum utility. \cite{mercuur2019value} shows these functions fulfill the stated requirements. By using (\ref{func:value}) and (\ref{func:utility}), we enable our agent to choose a demand following theories on values \cite{Schwartz2012}  and \cite{Cooper2011a} empirical results on the UG.
    
    \subsubsection{Norm-Based Agent} The norm-based agent uses the replies it observes from the other agent to determine what an agent (based on its norms) should demand. In \cite{mercuur2019value}, following theories by \cite{Ostrom2007} and \cite{fishbein2011predicting}, it is stated that a norm exists for a particular person when that person perceives what most other people do or expect it. \cite{mercuur2019value} provides a translation from this definition to a model for the UG. In the UG, the proposer cannot see what other proposers do, because the proposer is only paired with responders. Therefore, the proposer uses the responses of the responder to form an idea of what is expected (i.e., the norm). \cite{mercuur2019value} provides the following function to map the observed responses ($OR_{ak}$) seen by agent $a$ up to the current round $k$ to a demand $d \in \{0...P\} \subset \mathbb{Z}$, 
    \begin{equation}
    normDemand(OR_{ak}) = \frac{\min\limits_{d\in RD_{ak}} d + \max\limits_{d \in AD_{ak}} d} {2}
    \label{func:norm}
    \end{equation}
    where $RD_{ak}\subset OR_{ak}$ is the set of demands that the proposer $a$ has seen rejected and $AD_{ak} \subset OR_{ak}$ is the set of demands that the proposer $a$ has seen accepted. The function thus uses two indicators to estimate the norm: the minimum of the rejected demands ($RD_{ak}$) and the maximum of the accepted demands ($AD_{ak}$). The rejection of a given demand indicates that the demand is higher than expected, thus the norm is lower than the minimum rejected demand. The acceptance of a given demand indicates that the demand is lower than expected (or perfect), thus the norm is higher (or equal) to the accepted demand. Equation (\ref{func:norm}) determines the norm as the average of these two indicators.  By using (\ref{func:norm}), we enable our agent to choose a demand following \cite{Ostrom2007} and \cite{fishbein2011predicting} theories on norms.
    
    We extend the model provided by \cite{mercuur2019value} to specify normDemand($OR_{ak}$) for three edge cases: a case where there are no observed responses ($OR_{ak} = \emptyset$), a case with only observed rejections ($RD_{ak} = OR_{ak}$), and a case with only observed accepts ($AD_{ak} = OR_{ak}$): 
    \begin{itemize}
    \item \textbf{$OR_{ak} = \emptyset$} $\rightarrow{}$ normDemand($OR_{ak}$) is drawn from the normal distribution $N(561.8,128.9)$, which is the normal distribution human demands follow in the empirical dataset we use of\citep{Cooper2011a}.\footnote{The norm that is drawn from the normal distribution is not used as input for the norm in subsequent rounds (i.e., the agent does not memorize it).}
    \item \textbf{$RD_{ak} = OR_{ak}$} $\rightarrow{}$ the agent averages between the minimum reject and halve the pie ($(\min\limits_{d\in RD_{ak}} d +0.5P)/2$).
    \item \textbf{$AD_{ak} = OR_{ak}$} $\rightarrow{}$ the agent averages between the maximum accept and the full pie ($(\max\limits_{d \in AD_{ak}} d + P)/2)$).
    \end{itemize} 
    \noindent
    In \cite{mercuur2019value}, the agent determines a randomized demand as the norm in all these cases. By specifying these edge cases, we improve the SHA model aiming to better represent human behavior.
    
    \subsubsection{Agent with Values and Norms}
    This paper combines the value-based agent and the norm-based agent to model an agent that uses both values and norms. Values and norms are decision-making concepts the agent uses to decide what action is good and what action is bad \cite{miller2018explanation}.  \cite{dechesne2013no} used values as more abstract ideals and norms as the more concrete context-dependent rules that follow from these ideals. Norms follow from values, but when agents copy the norms (but not values) from other agents the two can conflict. For example, one copies working over-hours although this is in conflict with its own values. This paper presents a model where an agent uses a weight ($vw_a$) to combine the best action based on values and the best action based on norms into the best action based on both values and norms.
    
    For a proposer an action is defined as a demand $d \in \{0...P\} \subset \mathbb{Z}$.  The demand $d$ for an agent $a$ in round $k$ is determined by
    \begin{multline}
            \label{func:demand_weight}
            d_{ak}(di_a, vw_a, OR_{ak}) =  vw_a \times valueDemand (di_a) \\
            + (1-vw_a) \times normDemand (OR_{ak}),
    \end{multline}
    
    \noindent
    where $di_a$ stands for the difference in importance an agent attributes to wealth versus the importance it attributes to fairness, $vw_a \in [0, 1]$ stands for the weight an agent attributes to its own values (versus the weight it attributes to norms) and $OR_{ak}$ stands for the set of observed replies the agent has seen. Thus what an agent demands is the result of weighting the result of two functions described in \cite{mercuur2019value}: the $valueDemand(di_a)$ (\ref{func:value}) and the $normDemand(OR_{ak})$ (\ref{func:norm}).
    
    The above focuses on the demands of the SHA-proposer (SHA-P), but to simulate the UG we also need SHA-responders (SHA-R). The SHA-R is defined the same way as to the proposer model except that it determines a threshold $t\in [0,P]$ (instead of a demand $d$) and has to base its norms on a set of observed demands (instead of observed responses). The SHA-R uses the same function as the SHA-P (\ref{func:demand_weight}) to determine the threshold. If the demand is higher than the threshold the responder rejects it, if the demand is lower than or equal to the threshold accepts it. The responder determines a threshold appropriate for its values analogue to the proposer, that is, by using (\ref{func:value}) and (\ref{func:utility}). The responder determines the norm for the threshold by averaging over the observed demands (instead of the observed responses). The proposed demand is thus seen as a signal from the proposer that this is what the proposer considers `normal'.\footnote{To be exact, the proposed demand should be considered as what the proposer considers a `normal' threshold. If it considered a higher threshold to be normal it would have demanded less, if it considered a lower threshold normal it would have demanded more.} By using a threshold $t$ based on its values and the norm (from observed demands), the responder uses values and norms to reject or accept the proposed demand.
    
    Above we presented a model that uses both values and norms using two agent-specific variables: the difference in importance ($di$) and the value-norm weight ($vw$). The difference in importance specifies how much an agent values wealth over fairness and is normally distributed over agents $N(\mu_{di},\sigma_{di})$, being $\mu_{di}$ the average value and $\sigma_{di}$ the standard distribution of a normal distribution for $di$. The value-norm weight specifies how much an agent weighs values over norms and is normally distributed over agents $N(\mu_{vw},\sigma_{vw})$. Both variables are constant over the different rounds. However, the $normDemand(OR_{ak}$) differs per round as the observed replies vary. To reproduce the demands and responses humans give, the parameters $\mu_{di}$, $\sigma_{di}$, $\mu_{vw}$ and $\sigma_{vw}$ need to be calibrated to the right settings.
    
    \subsection{Calibrating the SHA to humans}
    We found that $\mu_{di} = 0.5$, $\sigma_{di} = 0.25$, $\mu_{vw} = -0.6$ and $\sigma_{vw} =1.14$ produced simulated demands and responses that are closest to the empirical data on human demands and responses extracted from the dataset provided by \citep{Cooper2011a}. This meta-study combines the data of 6 empirical studies to create a dataset of 5950 demands and replies made by humans in the same scenario as we describe in Section \ref{sec:ug}. We used five performance measures for which we compared the synthetic data on the SHA to the empirical data on real humans: average demand $\mu_d$, standard deviation in demand $\sigma_d$, average acceptance rate $\mu_a$, standard deviation in acceptance rate $\sigma_a$ and the standard deviation in the demand solely based on values $\sigma_{vd}$.\footnote{The standard deviation in the demand solely based on values $\sigma_{vd}$ was added to ensure agents vary in what values they find important ($di_a$). The $\sigma_{vd}$ for humans is postulated instead of extracted from empirical data.} The SHA is compared to humans on both round 1 and round 10. We used the following procedure to find the optimal parameter settings: 
    
    \begin{enumerate}
    \item Run simulations of the UG in Repast Simphony with different parameter settings ($\mu_{di} \in [-1,1]$ and $\mu_{vw} \in [0,1]$) and different random seeds ($r\in[1,30]$) and obtain the resulting demands and acceptance rate ($\mu_d$, $\sigma_d$, $\mu_a$, $\sigma_a$);
    \item For each run: average the performance measures with the same parameter settings, but different random seeds;
    \item For each parameter setting: calculate the normalized root mean square error (NRMSE) between the simulated results and the human results;
    \item Pick the parameter setting with a minimal NRMSE.
    \end{enumerate}
    Table \ref{tab:performancemeasures} compares the resulting demands and responses for the SHA (when NRMSE is minimized) and the empirical data on human results. We interpret these results as that our SHA can produce a distribution of demands and accepts that it is fairly close to that of humans (NRMSE = 11.0). The remainder of the paper uses these parameter settings to simulate human behavior based on values and norms.
    
    \setlength{\tabcolsep}{12pt}
    \begin{table}[]
    \centering
    \caption{A comparison of the distribution of demands and responses between empirical data on humans and synthetic data on simulated human agents (SHAs) given the parameter settings with the best fit ($\mu_{di} = 0.5$, $\sigma_{di} = 0.25$, $\mu_{vw} = -0.6$ and $\sigma_{vw} =1.14$).}
    \label{tab:performancemeasures}
    \resizebox{\textwidth}{!}{%
    \begin{tabular}{@{}ccccccc@{}}
    \toprule
    \rowcolor[HTML]{FFFFFF} 
     &  & \multicolumn{5}{c}{\cellcolor[HTML]{FFFFFF}Performance Measures} \\ \cmidrule(l){3-7} 
    \rowcolor[HTML]{FFFFFF} 
    Round & Source & \textbf{$\mu_d$} & \textbf{$\sigma_d$} & \textbf{$\mu_a$} & \textbf{$\sigma_a$} & \textbf{ $\sigma_{vd}$} \\ \midrule
    \rowcolor[HTML]{EFEFEF} 
    1 & empirical (human) & 561.8 & 128.9 & 0.806 & 0.40 & 128.9 \\
    \rowcolor[HTML]{FFFFFF} 
     & synthetic (SHA) & 557.9 & 91.1 & 0.876 & 0.29 & 109.2 \\
    \rowcolor[HTML]{EFEFEF} 
    10 & empirical (human) & 584.2 & 98.66 & 0.868 & 0.34 & 122.5 \\
    \rowcolor[HTML]{FFFFFF} 
     & synthetic (SHA) & 646.8 & 90.89 & 0.923 & 0.23 & 109.2 \\ \bottomrule
    \end{tabular}%
    }
    \end{table}

    \section{Estimating relative preferences on values and norms} 
    
\begin{figure}
        \centering
        \includegraphics[width=\textwidth]{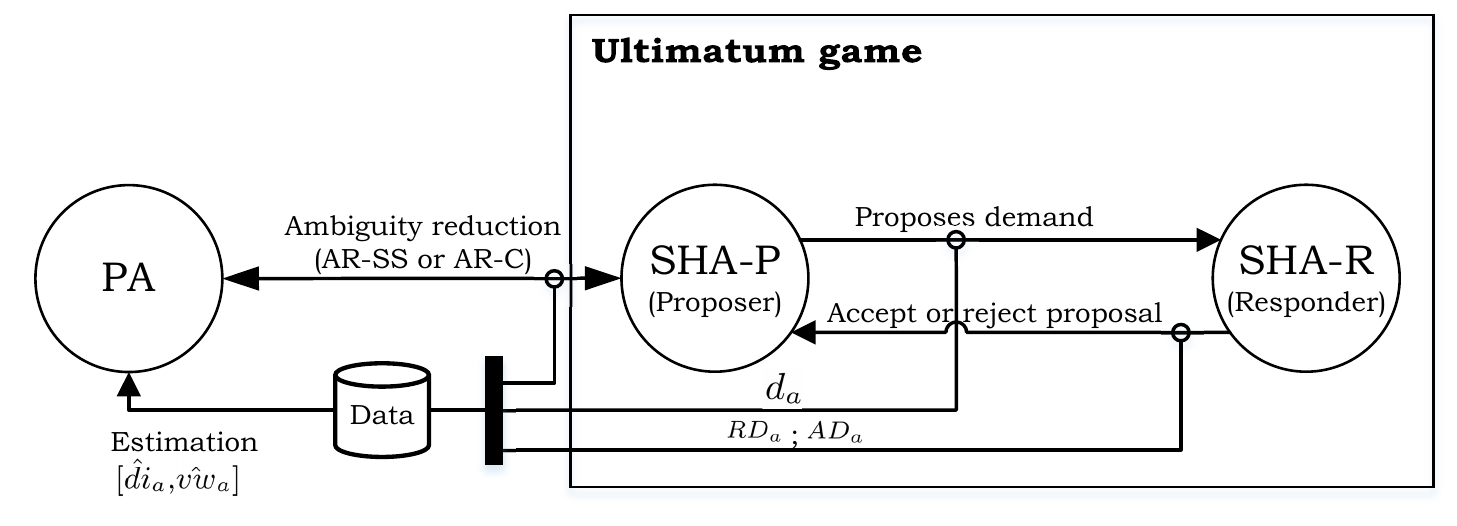}
        \caption{Conceptual model.}
        \label{fig:conceptual_model}
    \end{figure}

    The conceptual model of the estimation process is presented in Fig. \ref{fig:conceptual_model}. The \textit{Profiling Agent} (PA) is responsible to estimate the relative preference of a given SHA-P towards values and norms  ($[\hat{di}_a,\hat{vw}_a]$), and whenever necessary, interact with the SHA-P to reduce ambiguity in the estimation. 
    
    The problem of estimating values and norms in the context of this work is translated to the estimation of ${di}_a$ and ${vw}_a$, represented as [$\hat{{di}}_a$,$\hat{{vw}}_a$]. The estimation of relative preferences from behavior was assessed in different ways, such as via Inverse Reinforcement Learning (IRL) algorithms \cite{hadfield2017inverse,abbeel2004apprenticeship,levine2011nonlinear,2018occam} and learning a utility function with influence diagrams  \cite{nielsen2004learning}. In this work, we use the UG as a simple context (two values and one norm), aiming for clarity on understanding the conditions necessary to properly estimate the relative value and norm preference from behavior and, whenever necessary, how to reduce ambiguity.
    
    \vspace{1cm}
    
    In the context of the UG we can perform the estimation of the relative preferences by a exhaustive search process on the estimators, according to:
    
    \begin{equation}
    [\hat{di}_a,\hat{vw}_a] = \argmin_{di_a\in [-0.15;1.79]; vw_a \in [0;1]}   \dfrac{\sum_{k=1}^{m} \left| \hat{d}_{ak}({di}_a, {vw}_a, OR_{ak} ) - d_{ak}\right|}{k} 
    \label{func:argmax_estimation}
    \end{equation}
    
    
    \noindent 
    where $\hat{d_{ak}}$ is calculated by (\ref{func:demand_weight}), and $m$ represents the number of rounds from the UG analyzed in the estimation. In short, the estimators $\hat{{di}_a}$ and $\hat{{vw}_a}$ are defined as the preference set that minimizes the deviation between the observed demand ($d_{ak}$) and the demand that was calculated by an exhaustive search process ($\hat{d_{ak}}$). The range for $di_a$ which has an effect on the demand by the SHA-P is $[-0.15;1.79]$. With $vw_a \in [0;1]$, and using a step of 0.01 to explore both variables, 19,392 evaluations of the fitness function are necessary for each estimation process.
    
    \subsection{Reducing ambiguity} 
    
    Ambiguity arises whenever the number of elements of $\hat{{di}_a}$ and $\hat{{vw}_a}$ is greater than 1. The PA will try to reduce this ambiguity by taking the role of the \textit{responder} on the UG. This interaction might be interpreted as an elicitation process or, in simple, the PA will ask the SHA-P ``questions''. Different ways of ``asking these questions'' may produce higher or lower quality answers, and consequently a higher or lower change in the confidence on the estimation. We will explore two different approaches to how the PA interacts with the SHA-P. These interactions are a ``side-game'', i.e. changes in $OR_{a}$ will not, by definition, influence future actions of the SHA-P.
    
    
    \vspace{1cm}
    
    \subsubsection{Ambiguity Reduction by exploring the Search Space (AR-SS)}
    
    The approach to ambiguity reduction via exploring the search space ($AR-SS$) poses the following ``question'' to the SHA: \textit{What would your next demand be?}. Based on the demand (``answer'') provided by the SHA-P, the PA will reject or accept it to explore the search space (Algorithm \ref{algo_AR-SS}). This approach increases the search space, i.e. the distribution of $OR_{a}$ in the dataset, by rejecting demands lower than rejected before and accepting demands higher than accepted before.
    
    \begin{algorithm}[h]
    \SetKwData{Left}{left}\SetKwData{This}{this}\SetKwData{Up}{up}
    \SetKwFunction{Union}{Union}\SetKwFunction{FindCompress}{FindCompress}
    \SetKwRepeat{Do}{do}{while}
    \SetKwInOut{Input}{input}\SetKwInOut{Output}{output}
    \Input{$\hat{di}_a$, $\hat{vw}_a$, $max_{int}$,$RD_{a}$,$AD_{a}$,$OR_{a}$, $d_{a}$, $k$ }
    \Output{$\hat{di}_a$, $\hat{vw}_a$,$RD_{a}$,$AD_{a}$,$OR_{a}$, $count$}
    \BlankLine
    $n_{solutions} \leftarrow $number of solutions $([\hat{di}_a, \hat{vw}_a])$\\
    $k \leftarrow k+1$\\
    Calculate $OR_{ak}$ \\
    $count \leftarrow 0$\\
    
  \While{$n_{solutions}>1$ AND $count < max_{int}$}
        {
        Calculate $\hat{d}_{ak}(\hat{di}_a, \hat{vw}_a, OR_{ak})$ (\ref{func:demand_weight})\\
        Include $OR_{ak}$ and $\hat{d}_{ak}$ to the data set\\
        Estimate [$\hat{di}_a$,$\hat{vw}_a$]  (\ref{func:argmax_estimation})\\
        $n_{solutions} \leftarrow$ number of solutions $([\hat{di}_a, \hat{vw}_a])$\\
    
        \uIf{$d_{ak} < min({RD_{a}})$ OR $\nexists$ $RD_{a}$}
        {
         $RD_{ak} \leftarrow d_{ak}$ \\
        }
      \uElseIf{$d_{ak} > max({AD_{a}})$ OR $\nexists$ $AD_{a}$}
        {
         $AD_{ak} \leftarrow d_{ak}$\\
        }
      \BlankLine
        $k \leftarrow k+1$\\
        Calculate $OR_{a,k}$ \\
        $count \leftarrow count + 1$\\
        }
    \caption{Ambiguity Reduction by exploring the Search Space ($AR-SS$)}
    \label{algo_AR-SS}
    \end{algorithm}
    
    \subsubsection{Ambiguity Reduction via Counterfactuals (AR-C)}
    
    Counterfactuals are mental representations of alternatives to events that have already occurred \cite{roese1997counterfactual}, frequently represented by conditional propositions related to questions in the ``what if'' form. Philosophical discussion on counterfactuals has been present for ages, including works from David Hume and John Stuart Mill \cite{pearl2019seven}, and it is considered an intrinsic element of causality. People use counterfactuals often in daily life to create alternatives to reality guided by rational principles.
    
    Our approach to reduce ambiguity on the estimation of values and norms via counterfactual ($AR-C$) poses the following ``question'' to the SHA-P: \textit{What would your next demand be if your opponent had accepted instead of rejected your proposal on round ``x''?} or \textit{What would your next demand be if your opponent had rejected instead of accepting your proposal on round ``x''?}. As presented in Algorithm \ref{algo_AR-C}, the PA will `ask' the `question' related to the round that will lead to a broader search space in terms of the observed social norm ($OR_a$).
    
    \begin{algorithm}[h]
    \SetKwData{Left}{left}\SetKwData{This}{this}\SetKwData{Up}{up}
    \SetKwRepeat{Do}{do}{while}
    \SetKwInOut{Input}{input}\SetKwInOut{Output}{output}
    \Input{$\hat{di}_a$, $\hat{vw}_a$, $RD_{a}$,$AD_{a}$,$OR_{a}$, $d_{a}$,$k$ }
    \Output{$\hat{di}_a$, $\hat{vw}_a$,$RD_{a}$,$AD_{a}$,$OR_{a}$, $count$}
    
    \BlankLine
    $max_{int} \leftarrow k$ \\
    $count \leftarrow 0$\\
    $n_{solutions} \leftarrow $number of solutions $([\hat{di}_a, \hat{vw}_a])$\\
    
     \While{$n_{solutions}>1$ AND $count < max_{int}$}{
        \For{$i=1:k$}
        {
        \uIf{$\exists$ $AD_{ai}$ (Proposal was \textbf{accepted} in round $i$)}{
                \uIf{$d_{ai} < min({RD_{a}})$ OR $\nexists$ $RD_{a}$}{
                    $RD_{ai} \leftarrow d_{ai}$\\
            }
        }
         \uElseIf{$\exists$ $RD_{ai}$ (Proposal was \textbf{rejected} in round $i$)}{
            \uIf{$d_{ai} > max({AD_{a}})$ OR $\nexists$ $AD_{a}$}{
                $AD_{ai} \leftarrow d_{ai}$\\
            }
         }
        Calculate $OR_{ai}$ \\
         $score_i \leftarrow min (|OR_{ai} - OR_{a}|$)\\
        }
        $k \leftarrow k+1$\\
        $OR_{ak} \leftarrow OR_{az}$, where $z$ is the iteration where $score$ is maximum \\
        Calculate $\hat{d}_{ak}(\hat{di}_a, \hat{vw}_a, OR_{ak})$ (\ref{func:demand_weight})\\
        Include $OR_{ak}$ and $d_{new}$ to the data set\\
        Estimate [$\hat{di}_a$,$\hat{vw}_a$]  (\ref{func:argmax_estimation})\\
        $n_{solutions} \leftarrow $number of solutions $([\hat{di}_a, \hat{vw}_a])$\\
        $count \leftarrow count + 1$\\
     }
    
    \caption{Ambiguity Reduction via Counterfactuals (AR-C)}\label{algo_AR-C}
    \end{algorithm}
    
    \section{Results}
    
    To test the methods presented in this paper we analyzed the behavior of agents acting as proponents (SHA-P) in 100 runs of the UG (each with 20 rounds), in terms of precision and confidence. The first will be evaluated by the root mean squared error (RMSE) between the observed demand ($d$) and the estimated demand ($\hat{d}$), while the latter by the number of elements in $[\hat{di}_a, \hat{vw}_a]$. A small number of elements represent high confidence (low ambiguity), while a large number of solutions represent low confidence (high ambiguity). It is guaranteed the number of elements of $[\hat{di}_a, \hat{vw}_a]$ is greater than zero, since we perform an exhaustive search in the complete range of $di_a$ and $vw_a$. With these results, we aim to analyze the conditions under which the proposed methods can estimate preferences on values and norms in a precise and confident manner.
    
    \subsection{Estimation of values and norms}
     \label{sec:Est_ValuesNorms}
    
\begin{figure}
        \centering
        \includegraphics[width=\textwidth]{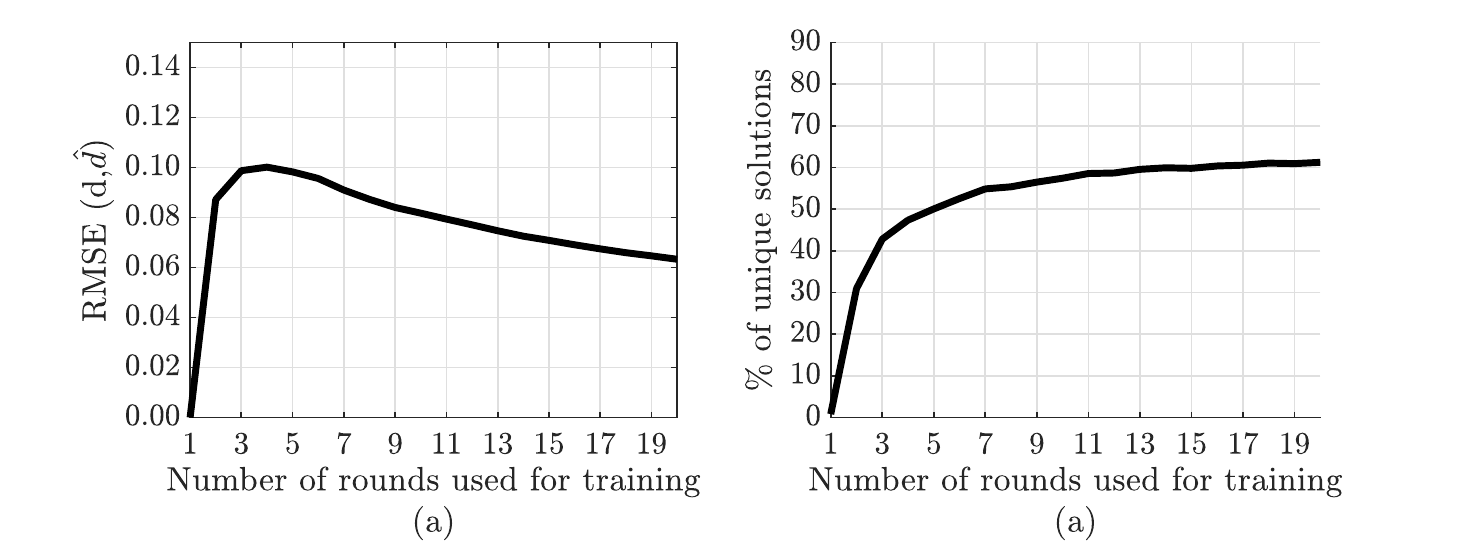}
        \caption{(a) Precision of the estimation process. (b) Percentage of unique solutions for the estimation process.}
        \label{fig:Results_training_utility}
    \end{figure}

     
    The precision in the estimation increased with the number of rounds used, given that an estimation is made by observing at least four rounds (Fig. \ref{fig:Results_training_utility}. (a)). Using less than four rounds the estimated values and norms $([\hat{di}_a, \hat{vw}_a])$ predicted a demand ($\hat{d}$) that is very close to the real demand\footnote{Given the deterministic model of the SHA, it might be expected that the RMSE should tend to zero. This is not the case because $[d$, $valueDemand$, $normDemand] \in  \mathbb{Z}$, and therefore a rounding operator is used.}, but most of these estimations will not be able to generalize among other contexts (i.e. other observed response - $OR_a$). 
    
    When considering between one and four rounds the estimations were ambiguous: on average 69.2 different sets of estimations led to the same demand  (Fig. \ref{fig:Results_training_utility}. (b)). Nevertheless, the confidence in the estimation increased with the number of rounds used. The steep curve reaching round four appears in both figures, showing a correlation between this initial precision in the estimation of the demand with the ambiguity on estimating preferences toward values and norms. Even with an increasing number of rounds observed, the percentual of unique solutions reached an average of only ca. 60\%. These different estimations may lead to undesired properties of an agent that wants to act on behalf of the true values and norms of a given person. 
    
    \subsection{Reducing ambiguity}
    
\begin{figure}
        \centering
        \includegraphics[width=\textwidth]{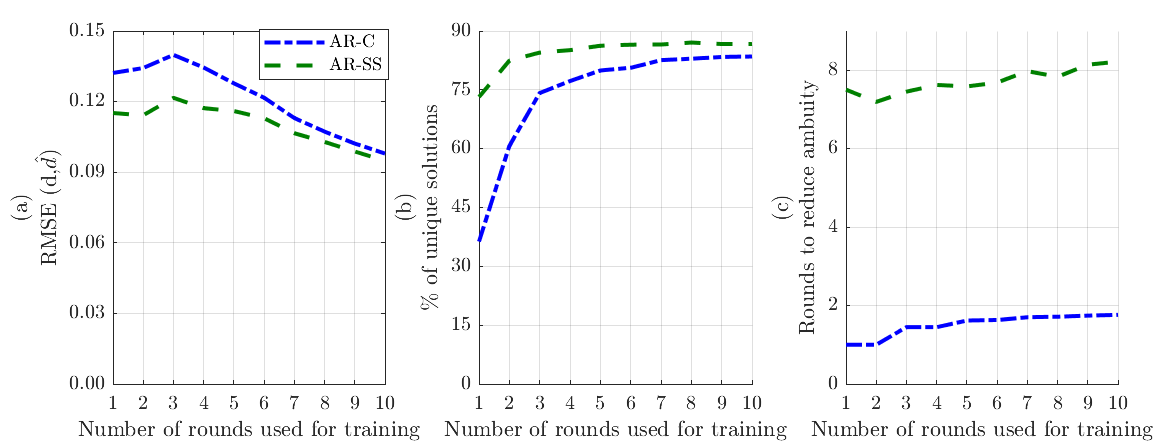}
        \caption{Reducing ambiguity: (a) Precision; (b) Unique solutions; (c) Number of interactions made for reducing ambiguity.}
        \label{fig:Results_RA_3rows}
    \end{figure}

    Both proposed methods were able to increase confidence in the estimation of preferences on values and norms $([\hat{di}_a, \hat{vw}_a])$. Comparing Fig. \ref{fig:Results_RA_3rows}. (a) with Fig. \ref{fig:Results_training_utility}. (a), the RMSE between the calculated and observed demand are slightly higher, but still may be considered adequate in absolute values, given that $d \in \{0 ... 1000\} \subset \mathbb{Z}$.
    
    The methods $AR-SS$ and $AR-C$ did not increase confidence in the same manner, differing in their applicability. While $AR-SS$ was able to reach estimations with less ambiguity than $AR-C$ (Fig. \ref{fig:Results_RA_3rows}. (b)), it required in average of 4.9 more interactions with the user (Fig. \ref{fig:Results_RA_3rows}. (c))\footnote{The x-axis in Fig. \ref{fig:Results_RA_3rows} relates to the number of rounds used during the initial estimation process, described in section \ref{sec:Est_ValuesNorms}. The additional number of interactions performed by each method to improve the confidence in the estimations is not included in the x-axis but presented in Fig. \ref{fig:Results_RA_3rows}. (c).}. In 33.2\% of the cases where the estimation was ambiguous, $AR-SS$ could provide a unique solution, while $AR-C$ provided it for 27.2\% of the cases. Considering both the confidence and the number of interactions needed, $AR-C$ increased the number of unique solutions by 16.5\% per interaction with the SHA-P, while $AR-SS$ increased it by 4.1\% per interaction.
    
    Table \ref{tab:discussion} summarizes the results when the estimation is performed using 10 rounds of observed data. We can see that precision varied only slightly, but the confidence in the estimation increased significantly when using any of the \textit{ambiguity reduction} methods proposed. The columns related to the ``Standard deviation'' on Table \ref{tab:discussion} provide parameters to understand the dispersion of the search space\footnote {in our case $OR_a$: given that preference to values and norms are constant, demand is defined according to \ref{func:demand_weight})} and of the estimations ($[\hat{di}_a, \hat{vw}_a]$). While the search space scatters with the ambiguity reduction methods, there is a reduction in the dispersion for the estimations. In other words, as the search space covered by $OR_a$ increases, estimations become less ambiguous (and more gathered).
    
    \setlength{\tabcolsep}{12pt}
    \begin{table}[]
    \centering
    \caption{Summary of the results for the estimation using 10 rounds, and for the subsequent ambiguity reduction methods.}
    \label{tab:discussion}
    \resizebox{\textwidth}{!}{%
    \begin{tabular}{ccccccc}
    \toprule
    \rowcolor[HTML]{FFFFFF} 
     &  & & &\multicolumn{3}{c}{\cellcolor[HTML]{FFFFFF}Standard deviation} \\ \cmidrule(l){5-7} 
    \rowcolor[HTML]{FFFFFF} 
    Method & Precision & \% Unique & Rounds on AR & \textbf{$OR_a$} & \textbf{$\hat{di}_a$} & \textbf{$\hat{vw}_a$}  \\ \midrule
    \rowcolor[HTML]{EFEFEF} 
    Estimation & 0.082  & 57.4 &  - & 36.1 & 0.021 & 0.008 \\
    \rowcolor[HTML]{FFFFFF} 
    AR-SS & 0.095 & 86.7 & 8.2 & 42.6 & 0.011 & 0.003 \\
    \rowcolor[HTML]{EFEFEF} 
    AR-C & 0.098 & 83.5 & 1.8 & 45.5 & 0.013 & 0.003 \\ \bottomrule
    \end{tabular}%
    }
    \end{table}
    
    \section{Discussion}
    
    AAs must be able to estimate human's relative preferences towards values and norms to align their behavior with them. This is not an easy task since a different set of preferences might lead to the same observed behavior (ambiguity problem). \cite{2018occam} demonstrated that `normative assumptions' are needed to estimate an agent's reward function (in our case, values and norms). In this work, we proposed an ABM that uses both values and norms to account for these `normative assumptions', and methods to estimate the SHA preferences and reduce the ambiguity in these estimations.
    
    The first contribution, the proposed ABM, was built on previous works \cite{malle2006mind,miller2018explanation}, and especially \cite{mercuur2019value}. We extended the ABM works to use \textit{both} values and norms for determining the actions of the agents. The model was calibrated to represent human behavior from empirical data. Future works can improve this model by specifying in more detail the use of norms and values, considering other values than wealth and fairness, constraining behavior by using deontic operators \cite{mercuur2019value}, and incorporating models that represent human bounded rationality. Furthermore, different application scenarios with a more complex view on value tensions e.g. non-zero sum game problems considering privacy and security can be modeled.
    
    The second contribution of this work was focused on understanding to what extent the proposed methods were able to estimate the relative importance attributed to values and norms by a given agent. We show that even when considering `normative' assumptions, represented by the PA's knowledge of the decision-making process of the SHA in a relatively simple context (the UG), ambiguity may still be present on estimating preferences for value alignment. To ignore this ambiguity might lead to great regret and unethical actions. 
    
    In the experiment performed we observed two general conditions under which an estimation of the preferences was possible, namely: heterogeneity on the observations and a sufficient number of observations. When observing at least four rounds of the UG, only ~50 to ~60\% of the estimations were unambiguous. We proposed two methods for reducing ambiguity: one via exploring the search space ($AR-SS$) and one using counterfactuals ($AR-C$). The first approach, $AR-SS$,  interacts with the SHA-P considering the last observed norm ($OR_{ak}$), which might be suitable for interacting with people in the real world due to the influence of short-term memory on decision making \cite{del2013multifold}. The latter approach, $AR-C$, uses counterfactuals, which are a part of causal reasoning. The results presented in the previous section show that targeting ``imaginative'' scenarios related to a specific round of the UG significantly increase the efficiency of the process. Nevertheless, it might be very demanding for a person to be able to go through this thought imaginary process and remind the precise social norm at a specific point in time.  We can conclude that the $AR-SS$ method is more suitable where there are no restrictions regarding the number of interactions with the user, while $AR-C$ can improve confidence in situations where a limited number of interactions is desired. 
    
    Both methods act with the assumption that the only way the PA can interact with the SHA is by taking the role of the responder in the UG. Going beyond this assumption, we also evaluated a different approach: the PA can directly define a value for the social norm ($OR_{ak}$). If we consider $OR_{a} \in [0;1000]$, the ambiguity was almost eliminated:  ~97\% of unique solutions on average, considering up to 20 interactions. If we consider a more realistic assumption $OR_{ak} \in [500;1000]$ the results were, in terms of the percentage of unique solutions, between the levels found by $AR-SS$ and $AR-C$ when using less than 6 rounds for training, and slightly better than $AR-SS$ when using 6 or more rounds. Future works can evaluate this hypothesis. We suggest that such experiments be done either considering improved models of human memory and cognition process or in laboratory settings.
    
    The limitations of this works include the impossibility of directly generalizing the findings and methods to other contexts, the assumption that values are stable, and the lack of testing of the approach with humans in realistic settings as well as in more complex settings. 
    
    \section{Conclusion}
    
    This paper aimed to investigate to what extent an AA might be able to estimate the relative preferences attribute to human values and norms, including methods to reduce ambiguity. Insight into the use of models to support the estimation of values and norms was obtained during the discussions, mainly the need for heterogeneity on the observations and also ways to reduce ambiguity. Especially the use of counterfactuals via the $AR-C$ approach showed it can be of great value in terms of a trade-off between increasing efficiency and avoiding excessive interactions/questions with humans. We showed that even in a simple context, considering models that represent values and norms (`normative assumptions'), and using a exhaustive search process, ambiguity cannot be easily be avoided in estimation of preference on values and norms. To ignore this ambiguity might lead to great regret and misalignment between machine behavior and human values and norms.
    
    \subsection*{Acknowledgements}
    This work was supported by the AiTech initiative of the Delft University of Technology.
    
    \bibliography{bibliography_paper}
    \bibliographystyle{splncs04}
	
\end{document}